# Towards Design and Development of an ArUco Markers-Based Quantitative Surface Tactile Sensor

Ozdemir Can Kara, *Student Member, IEEE*, Charles Everson, and Farshid Alambeigi, *Member, IEEE*

*Abstract*— In this paper, with the goal of quantifying the qualitative image outputs of a Vision-based Tactile Sensor (VTS), we present the design, fabrication, and characterization of a novel Quantitative Surface Tactile Sensor (called QS-TS). QS-TS directly estimates the sensor's gel layer deformation in real-time enabling safe and autonomous tactile manipulation and servoing of delicate objects using robotic manipulators. The core of the proposed sensor is the utilization of miniature 1.5 mm × 1.5 mm synthetic square markers with inner binary patterns and a broad black border, called ArUco Markers. Each ArUco marker can provide real-time camera pose estimation that, in our design, is used as a quantitative measure for obtaining deformation of the QS-TS gel layer. Moreover, thanks to the use of ArUco markers, we propose a unique fabrication procedure that mitigates various challenges associated with the fabrication of the existing marker-based VTSs and offers an intuitive and less-arduous method for the construction of the VTS. Remarkably, the proposed fabrication facilitates the integration and adherence of markers with the gel layer to robustly and reliably obtain a quantitative measure of deformation in real-time regardless of the orientation of ArUco Markers. The performance and efficacy of the proposed QS-TS in estimating the deformation of the sensor's gel layer were experimentally evaluated and verified. Results demonstrate the phenomenal performance of the QS-TS in estimating the deformation of the gel layer with a relative error of <5%.

## I. INTRODUCTION

Tactile Sensing (TS) is the process of perceiving the physical properties of an object through a cutaneous touch-based interaction [1]. The acquired information can be very beneficial in many areas of robotics in which a robot interacts with hard or deformable objects. Examples include robotic manipulation (e.g., [2]–[5]), object texture or stiffness recognition [6], [7], slip detection [8], and human-robot interaction [9]. As the application of TS in robotics is increasing, there is an imminent need for the development of high-resolution and high-accuracy TS devices that can safely and robustly interact with a soft or rigid environment for dexterous and safe manipulation tasks.

Vision-based Tactile Sensors (VTSs) have recently been developed to enhance tactile perception via high-resolution visual information [10]–[13]. Particularly, VTSs can provide *qualitative* 3D visual image reconstruction and localization of the interacting rigid or deformable objects during robotic

*Research reported in this publication was supported by the National Cancer Institute of the National Institutes of Health under Award Number R21CA280747.

All authors are with the Walker Department of Mechanical Engineering, University of Texas at Austin, TX, USA Email: ozdemirckara@utexas.edu, charles.everson@utexas.edu and farshid.alambeigi@austin.utexas.edu

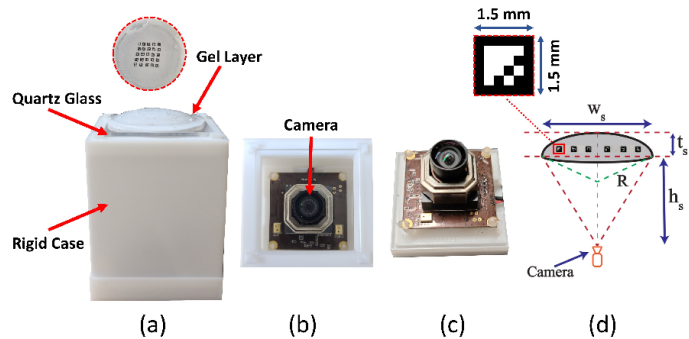

Fig. 1: The proposed Quantitative Surface Tactile Sensor (QS-TS) based on ArUco Markers. QS-TS consists of a deformable gel layer, a rigid case, Quartz transparent glass, and a camera: (a) Side view, (b) Top view, (c) Inner camera placement, and (d) Conceptual illustration of the QS-TS dimensions.

manipulation tasks by capturing very small deformations of an elastic gel layer that directly interacts with the objects' surface [14]. The evolution of digital and small size cameras has improved the fabrication, sensitivity, and integration of the VTSs with different robotic systems [15]. Further, advancements in fabrication methods, computer vision and machine learning algorithms have enabled the acquisition, real-time post-processing, and analysis of the high-resolution TS information provided by this sensor [16]–[19]. GelSight [16] is the most well-known VTS and has been used in various applications such as the surface texture recognition [20], geometry measurement with deep learning algorithms [21], [22], localization and manipulation of small objects [23], hardness estimation [24], and robotic manipulation [25]–[29]. Despite these advancements and as one of the major drawbacks of these sensors, most of the current VTSs still can solely provide *qualitative* visual deformation images and cannot yet directly measure a *quantitative* gel layer deformation and typically use machine learning and computer vision algorithm to estimate this deformation [21], [30], [31]. Moreover, a review of the literature shows that most of the performed ML-based estimation analyses either are limited to objects with simplified geometries or even ignore verifying the obtained estimated deformations with a ground truth value.

To address this limitation, various techniques have been proposed in the literature to provide a quantitative measure of the deformation when a VTS interacts with an object. These approaches can be mainly divided into two classes: marker-tracking-based and reflective-membrane-based sensors. For instance, GelForce [32], and Chromatouch [33], are examples of marker-tracking-based sensors, in which a pattern of markers is utilized within the elastomer body. When the interaction occurs between the sensing gel layer and the object, the markers' pattern is affected, and their movements can be processed to infer the tactile information. However, marker-based designs require an arduous and complex manufacturing procedure to robustly adhere and integrate the markers with the VTS gel layer. Examples of these procedures include casting or 3D-printing of gel layers [14] as well as relatively easier and inaccurate marker-printing approaches based on transfer papers or hand-writing [21], [24]. Nevertheless, these techniques may suffer from wrinkling or inconsistencies in the printing pattern, which directly deteriorates the accuracy of the deformation measurements [21], [24]. Kim et al. proposed an UVtac marker-based sensor based on their previous study [34] that solved the inconsistencies in the aforementioned printing method by utilizing the direct UV marker printing on the sensor surface with the expensive equipment, and comparably higher steps of fabrication. This approach, however, only provides the normal and shear force estimation as a quantitative evaluation in where the large estimation errors were seen at lower normal forces [35]. Moreover, Zhang et al. utilized optical flow to estimate the deformation of a dense color pattern and employed the transfer printing technique, and resolved the inconsistencies. The deformation estimation results of DelTact, however, suffered from the artifacts caused by the estimation algorithms [36].

On the contrary, reflective membrane-based sensors, called retrographic sensors (e.g., GelSight [16]), are employed for sensing the shape and texture of the objects through the analysis of the intensity change of the reflected light from the reflective elastic sensing surfaces [14]. Despite the benefits of using this technique, the flat surface of the gel layer can limit the sensing of the object's orientation. Moreover, tactile information regarding tangential forces and deformation cannot be accurately detected in this approach, and objects without textures or edges are not recognized very well in reflective membrane-based sensors [35]. To address the challenges associated with these two types of VTSs, Nozu et al. [37] developed a tactile sensor incorporating both reflective membrane-based sensing and marker-based design for in-hand object localization and force sensing. However, in this approach, utilized markers on the surface of the gel layer may occlude the view of the camera looking towards the reflective membrane, and therefore, deteriorating the quality of the extracted tactile information from the visual output. To mitigate this issue, recently, Kim et al. [35] proposed a tactile sensor that merges both reflective membrane-based sensing and marker-based design to decouple the marker and reflective membrane images and offer a 3-axis force estimation and object localization. Similar to the previous VTS, however, this sensor only provides an estimated interaction force between the VTS and the object and does not provide direct quantitative information on the deformation of the gel layer. Recently, Lin et al. [38] has also developed a VTS capable of measuring the 3D geometry of interacting objects and obtaining pose estimation. An iterative closest point algorithm was utilized to predict the pose that cannot yield real-time direct quantitative deformation information. A detailed review of similar technologies can also be found in [39].

As our main contributions and towards collectively addressing the above-mentioned limitations of existing VTSs in quantitative textural measurements of gel layer deformation and the learning-based methods (e.g., [24], [36]), in this paper, we present the design, fabrication, and characterization of a novel Quantitative Surface Tactile Sensor (QS-TS). This sensor can be integrated with a robotic hand to enable safe and autonomous tactile manipulation and servoing of delicate objects. The core of the proposed sensor is the utilization of miniature 1.5 mm × 1.5 mm synthetic square markers with inner binary patterns and a broad black border called ArUco Markers [40]. Each ArUco marker can provide real-time camera pose estimation that, in our design, is used as a quantitative measure for obtaining deformation of the QS-TS gel layer. Moreover, thanks to the use of ArUco markers, we propose a novel, consistent, easy, and intuitive fabrication procedure that mitigates the challenges mentioned above during the fabrication of VTSs and offers reproducible tactile sensors. Particularly, the proposed fabrication facilitates the integration and adherence of markers with the gel layer to robustly and reliably obtain a quantitative measure of deformation in real-time with an approximate error of <5%. This precise quantitative evaluation of the deformation would enable safe manipulation of objects with robotic hands equipped with QS-TS, similar to the studies performed in [23], [25]–[29].

## II. METHODOLOGY

In this section, we briefly describe the working principle of the proposed QS-TS based on ArUco markers and then explain in detail the fabrication procedure and deformation estimation algorithm of the QS-TS gel layer.

### A. ArUco Markers

Pose estimation is a computer vision problem that determines the orientation and position of the camera with respect to a given object and has great importance in many computer vision applications ranging from surgical robotics [41], augmented reality [42] to robot localization [43]. Binary square fiducial markers have been emerged as an accurate and reliable solution for the pose estimation problem in various robotic applications (e.g., autonomous robot manipulation [44], [45]). These markers offer easily discernible patterns with strong visual characteristics through their four corners and specific ID to obtain the camera pose. Moreover, their unique inner binary codes add robustness for the misdetections and reduce false positive detections [46]. Various fiducial marker libraries have been developed progressively in the literature to address the pose estimation problem such as AprilTag [47], ARTag [48], ArUco [40]. A detailed review of fiducial marker packages and descriptions can be found in [42]. In this work, we propose to use ArUco markers [40] for estimating the deformation of the VTS gel layer, as they can provide a high detection rate in real-time and allow the use of reconfigurable libraries with less computing time [42].

*B. QS-TS Design and Fabrication*

*1) Working Principle and Constructing Elements of QS-TS:* As demonstrated in Fig. 1 and similar to the GelSight sensor [21], QS-TS consists of (i) a dome-shape deformable silicone gel layer integrated with multiple ArUco markers for the quantification of the deformation field of the elastomer surface that directly interacts with an object, (ii) an autofocus camera that is fixed to the 3D printed frame of the sensor and faces toward the elastic gel layer to record the deformation of the gel layer and the movements of the ArUco markers, and (ii) a highly transparent Quartz glass layer (7784N13, McMaster-Carr) that supports the gel layer while providing a clear view to the camera. Of note, unlike the typical Gelsight sensors, QS-TS does not require Red, Blue, and Green (RGB) LEDs, as they create a glare on the inked surface of the printed ArUco markers preventing their edges from properly being detected during deformation. Instead, ambient lighting is preferred and used to not interfere with the ArUco markers edge detection. The *working principle* of QS-TS, similar to the GelSight sensor [21], is very simple yet highly intuitive, in which the deformation caused by the interaction of the silicone gel layer with the object can be captured by the autofocus camera and quantified through ArUco markers adhered to the surface of the gel layer and continuously moving with that.

*2) Fabrication Procedure of the QS-TS:* To fabricate the QS-TS, we used a 13 MP autofocus USB camera (IEights 4K 1/3 inch with an IMX 415 sensor and 5 - 50 mm varifocal lens) that was fixed to the rigid frame printed with a 3D printer (E2, Raise3D), and the PLA filament (Art White Extreme Smooth Surface Quality, Raise3D). An autofocus camera was a suitable selection for QS-TS as we could control and optimize the focal distance through the deformation procedure of the gel layer and always ensure a clear output image. In other words, a fixed focal length camera may create blurry visuals after exceeding the focus threshold and throughout the gel layer deformation. The rigid frame height was designed as $h_s$=55 mm based on the camera focus and 100° field of view. Of note, the zoom parameter of the autofocus camera and the distance between the camera and the ArUco markers were optimized to find the balance between the detection rate and the correct pose estimation of markers. Moreover, based on the design requirements, the QS-TS components and size can be readily optimized and scaled.

*Fabrication Procedure of the Gel Layer:* The following sections describe and compare the steps taken for fabricating a typical GelSight sensor and our novel fabrication method for the QS-TS sensor:

**GelSight Sensor:** <u>STEP I:</u> To fabricate the deformable gel layer (as illustrated in Fig. 1), we used a soft transparent platinum cure two-part (consisting of Part A and Part B) silicone (P- 565, Silicones Inc.), with a 14:10:4 ratio (A:B:C), in which Part C represents the phenyl trimethicone- softener (LC1550, Lotioncrafter). In this mixture, Part B functions as the activator of the two-part silicone, which can adjust the hardness of the silicone. Before pouring the silicone mixture into the hemispherical-shape silicone mold (Baker Depot mold for chocolate with a diameter of 35 mm), the surface of the silicone mold was coated with Ease 200 (Mann Technologies) twice to prevent adhesion and ensure a high surface quality after molding. After waiting for the drying of the coating for 10-12 minutes, the silicone mixture was poured into a silicone mold and then degassed in a vacuum chamber to remove the bubbles trapped within the mixture. Next, samples were solidified in a curing station (Formlabs Form Cure Curing Chamber). Of note, as demonstrated in Fig. 1 (d), the fabricated gel layer had width ($w_s$) and thickness ($t_s$) of 33.8 mm and 4.5 mm, respectively.

<u>STEP II:</u> After the curing step, black marker dots on the sensor surface could be attached either using waterslide decal paper [21] or manually marking by hand [24]. For the first option, the marker dot pattern was printed on the glossy side of the water transfer paper via a laserjet printer. Then, the transfer paper was soaked in medium-temperature water to wet the paper surface to maneuver and peel the printed side off easily. Afterward, the transfer paper was placed on the dome-shape gel layer with the marker dots facing up while separating the backing paper. Of note, this arduous procedure demands multiple repetitions and requires experience in working with the decal papers for the integration of it on the sensor surface. Even if the transfer paper is placed correctly on the surface of the gel layer, it will most likely be wrinkled when it interacts with an object, and therefore deteriorates the sensor sensitivity and quality of the output images. The aforementioned studies' approaches are relatively manageable but do not provide a direct and accurate quantitative measure for the gel layer deformation, that is one of the main challenges in the robotic manipulation of different objects.

<u>STEP III:</u> This step includes covering the printed markers on the gel layer's surface. To this end, first, the matte-colored aluminum powder (AL-101, Atlantic Equipment Engineers) was brushed on the gel layer's dome surface to avoid light leakage. Finally, a thin layer of silicone with the addition of grey pigment (a blend of both black and white pigments- Silc Pig Black, Silc Pig White, Smooth-On Inc) was poured, with identical proportion described in STEP I, on the surface of gel layer to stabilize the aluminum powder layer and prevent light leakage since there exist RGB LEDs within the rigid casing. Notably, the hardness of the gel layer sample was measured as 00-20 using a Shore 00 scale durometer (Model 1600 Dial Shore 00, Rex Gauge Company).

**QS-TS:** To fabricate the deformable gel layer for our novel sensor, we exactly followed the above-described procedure in STEP I. The significant change in the proposed QS-TS fabrication procedure occurs in *STEP II* and *STEP III* in which, instead of utilizing black dot marker patterns, we used 25 square ArUco Markers with the size of 1.5 mm × 1.5 mm and adhered them separately and one by one to the QS-TS gel layer surface. Each ArUco Markers were printed on a water transfer paper (Sunnyscopa) using a laserjet printer (Color Laser Jet Pro MFP M281fdw, Hewlett-Packard) with 600 × 600 dots per inch (DPI) to obtain the best printing quality from the utilized printer. Of note, the 1.5 mm × 1.5 mm marker size was determined after performing a few preliminary tests with the detection algorithm. It is worth mentioning that high DPI printing quality would enable using smaller marker sizes while still having a high detection rate. Before placing each ArUco marker, the sensor's surface was brushed with the versatile decal setting solution, Micro Set (Microscale Industries), to increase the adhesion and prepare the surface for the application of the transfer paper. After 5-10 minutes, each marker was placed with precision tweezers one by one by following the instructional procedures of the transfer paper to create a 5 × 5 array on the sensor surface. Of note, as each marker was positioned separately with an average of 2.5 mm distance, and ArUco Markers could be detected independently, regardless of their positioning, orientation, and uniformity, we did not face the previous problems in *STEP II* of GelSight preparation. In other words, thanks to the above-mentioned feature of the ArUco Markers, QS-TS can compensate for the misalignment of the markers in its sensing.

It is worth noting that due to the use of ambient light instead of LEDs, the proposed fabrication method eliminates the need for additional aluminum powder brushing in *STEP III*. Thus, as the last fabrication step, a thin layer of silicone mixture with the addition of white pigment (Silc Pig White, Smooth-On Inc) and the same proportion as in *STEP I*, was poured on the sensor surface to cover the ArUco Markers. Of note, white pigment is preferred to easily distinguish the black-colored patterns of the markers from the white background and aid the computer vision algorithm during the detection procedure.

*C. ArUco Marker Detection and Pose Estimation*

As demonstrated in Fig. 2, each ArUco Marker has its own binary codification and identification to provide a 3D position and orientation of the camera toward them. These fiducial markers have libraries based on OpenCV and are written in C++ [40]. This architecture employs square markers, which can be built for different dictionaries varying in number of bits and sizes. ArUco allows us to use reconfigurable predefined marker dictionaries, *DICT_X×X_Y*, in which X (4, 5, or 6) and Y (50, 100, 250, 1000) represent marker size in bits, and the number of the markers stored in this library, respectively[1]. The number of bits affects the confusion rate and the required camera accuracy and resolution. If the bit size is small, the patterns are more straightforward, and markers can be detected at lower resolution with the trade-off of a higher confusion rate. In addition to the bit size and number of markers in the dictionary, the inter-marker distance, the minimum distance between two separate fiducial markers, is a significant factor that can determine the error detection and correction capabilities. Mainly, larger markers and smaller dictionary sizes can also decrease the confusion between markers and aid in identifying a specific marker with higher accuracy. On the other hand, detecting markers with higher bit sizes becomes complex due to the requirement of a higher number of bits extracted from the image.

[1] https://docs.opencv.org/4.x/d5/dae/tutorial_aruco_detection.html

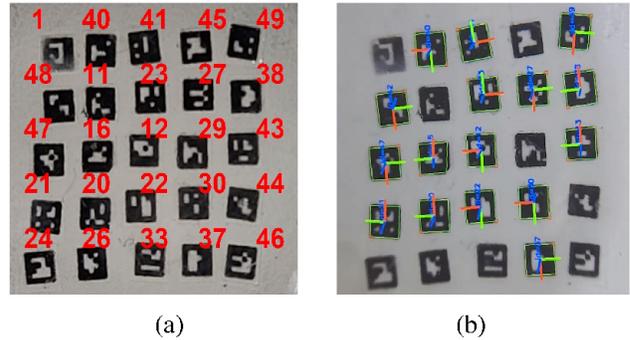

(a) (b)

Fig. 2: ArUco Markers integrated with the elastic gel layer of QS-TS. Each marker has its own ID to be recognized through the computer vision algorithm. (a) Zoomed view of the AruCo Markers with their unique identification numbers, (b) Randomly selected frame showing the detected ArUco markers.

In this study, we selected a 4×4 bit library with 50 ArUco Markers to have a robust detection of errors for our 25 ArUco markers (as demonstrated in Fig. 2), in consequence of the clarifications mentioned earlier. 5×5 array of markers, sizing 1.5 mm × 1.5 mm square, are prepared in order to create a balance between the total number of markers and the resolution and sensitivity of the sensor since a smaller size of markers means that more markers can be attached to the surface. All markers are generated through an online generator website [49]. Notably, the number, size, and attachment pattern of ArUco markers can be readily optimized and varied based on the application.

After integrating each 25 ArUco Marker on the deformable gel layer (shown in Fig. 2), we followed the Hamming coding algorithm proposed in [40], to optimize the low false negative rate for the pose estimation. The detection process started with the acquisition of the images from the autofocus camera. Then, these images were converted to grayscale to reduce the computational requirement and simplify the overall algorithm. Afterwards, contours were extracted as rectangles and filtered to obtain marker candidates, and perspectives were removed. Finally, the ID of each detected marker was generated with the rotation and translation vectors through the extraction of the unique binary codes secured in the markers and comparison of these codes with the selected marker dictionary. We implemented a Python solution for this algorithm to work in real-time while saving both detection rates and pose estimations to an Excel file. Figure 2 shows ArUco Markers placed on the elastic gel layer of QS-TS. As shown, each marker has its own ID to be recognized through the computer vision algorithm.

## III. EXPERIMENTAL SETUP AND PROCEDURE

Figure 3 demonstrates the experimental setup used to conduct characterization tests for QS-TS and obtain the displacement and orientation of each ArUco marker during the interaction with a flat object normally pressed on the gel layer. As shown, the experimental setup consists of the QS-TS, a 3D printed flat square object designed for testing the QS-TS deformation measurement, a single-row linear stage with 1 $\mu m$ precision (M-UMR12.40, Newport) for the precise control of the flat square plate displacement, a digital force gauge with 0.02 N resolution (Mark-10 Series 5, Mark-10 Corporation) attached to the linear stage to track the interaction force between the flat square plate and QS-TS, and a Dell Latitude 5400 for streaming and recording the video for data processing. We also utilized Spyder, the Scientific Python Development Software, to complete the camera calibration and process the acquired ArUco Marker data for the pose estimation.

In order to evaluate the performance of the QS-TS in detecting the ArUco markers attached to the sensor surface and the pose estimation, we first performed the camera calibration using OpenCV[2]. Of note, camera calibration was one of the most essential steps to identify markers correctly and determine each marker's accurate orientation and distance vectors- relying on $10 \times 7$ sized checkerboard[3] that was patterned with 1.5 mm $\times$ 1.5 mm squares. We utilized a set of 36 checkerboard images that were captured from different orientations and distances. After processing these checkerboard visuals, we obtained $3 \times 3$ camera intrinsic matrix and radial and tangential distortion coefficients. Notably, camera intrinsic matrix (CIM) is a unique matrix specific to a camera, consisting of both focal length ($f_x$,$f_y$) and optical centers ($c_x$ and $c_y$) [50]. It is expressed as a $3 \times 3$ matrix CIM = $[f_x\ 0\ c_x; 0\ f_y\ c_y; 0\ 0\ 1]$.

After the acquisition of both the CIM and distortion coefficients as an *.yaml* file, 3D printed flat square plate was attached to the force gauge using the threaded connection at the base. Then, QS-TS was fixed to the optical table to prevent any undesired slipping or sliding. Next, the linear stage was precisely moved until the square plate contacted the QS-TS. Notably, a force gauge was placed on the linear stage to detect the initial touch between the square plate and QS-TS to ensure that there was no deformation during the initial positioning. After arranging the hardware for measurements, the main detection and pose estimation algorithm was initialized to screen frame numbers, recognized marker IDs in real-time, and record numerical data involving detection rate, pose, and the orientation of each ArUco marker to the Excel file.

As shown in Fig. 3, during the characterization experiments and using linear stage, the flat square plate was pushed on the sensor's surface to introduce a maximum displacement of 2 mm for the sensor's gel layer. During this process and at each 400 $\mu m$ displacement, including the initial state (i.e., zero displacement), we recorded 200 image frames for each of the used 25 ArUco markers (shown in Fig. 2). Therefore, in total 1200 image frames were obtained for each ArUco marker. Next, at each of the six reading sequences, these frames were processed for each marker to calculate the average estimated position and their corresponding errors (as illustrated in Fig. 4). Of note, for repeatability purposes, this characterization experiment was repeated three times, and all the required values were collected and analyzed based on these repeated experiments.

[2]https://docs.opencv.org/4.x/dc/dbb/tutorial_py_calibration.html
[3]https://calib.io/pages/camera-calibration-pattern-generator

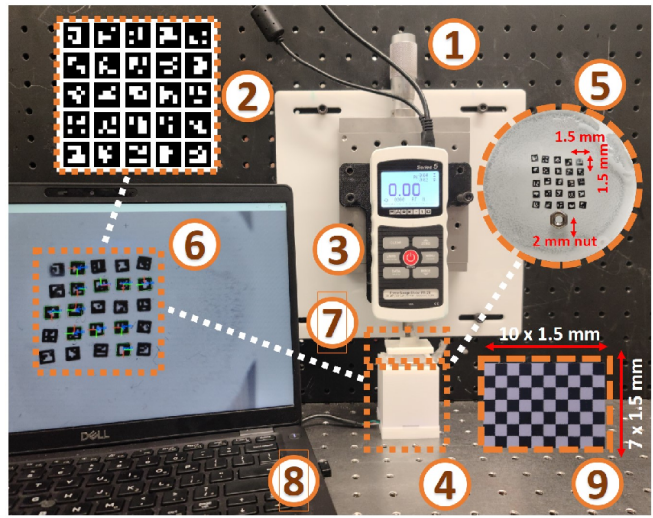

Fig. 3: The utilized experimental setup including: ①- M-UMR12.40 Precision Linear Stage, ②- 25 ArUco Markers pattern ③- Mark-10 Series 5 Digital Force Gauge, ④- Our proposed QS-TS sensor, ⑤- QS-TS's deformable gel layer with attached ArUco Markers. 2 mm nut were placed to indicate the small-scaled markers (1.5 mm $\times$ 1.5 mm), ⑥- Output of the real-time detection of each markers, ⑦- 3D printed flat object used for characterization experiments, ⑧- Dell Latitude 5400 laptop used for the data processing and real-time marker detection and pose estimation, ⑨- a $10 \times 7$ checkerboard used for the camera calibration with 1.5 mm $\times$ 1.5 mm squares.

## IV. RESULTS AND DISCUSSION

Figure 4 depicts the comparison of Z depth estimation of four exemplary ArUco Markers (i.e., ID 20, ID 21, ID 40, and ID 47 as marked in Fig.2) with their actual Z displacement applied using the linear stage. As shown in this figure, a total deformation of 2 mm with 0.4 mm intervals has been considered for analyzing each ID and evaluating the performance of the detection algorithm. This figure also reports the percentage of the relative error in estimating the deformation of the gel layer at the location of the considered IDs and during the deformation intervals. It can be easily seen from the error bars that a maximum error of 4.23% between the estimation and actual distance from the camera occurs for the ID 47 marker during all the intervals. On the other hand, the average measurement error for all other exemplary markers is around 2%, which signifies that the QS-TS sensor can reliably quantify the displacement of an object in different interaction locations with the sensor gel layer. Moreover, Fig. 5 shows the trajectory of four exemplary tags (i.e., ID 20, ID 21, ID 22, and ID 23 as marked in Fig. 2) the flat plate was pushed towards the Z direction with 0.4 mm intervals. As can be observed in this figure, the detection algorithm can reliably follow the trajectory of the detected markers during the deformation process. Notably, this critical feature enables shape reconstruction of the deformed gel layer to represent a dynamic deformation over time quantitatively.

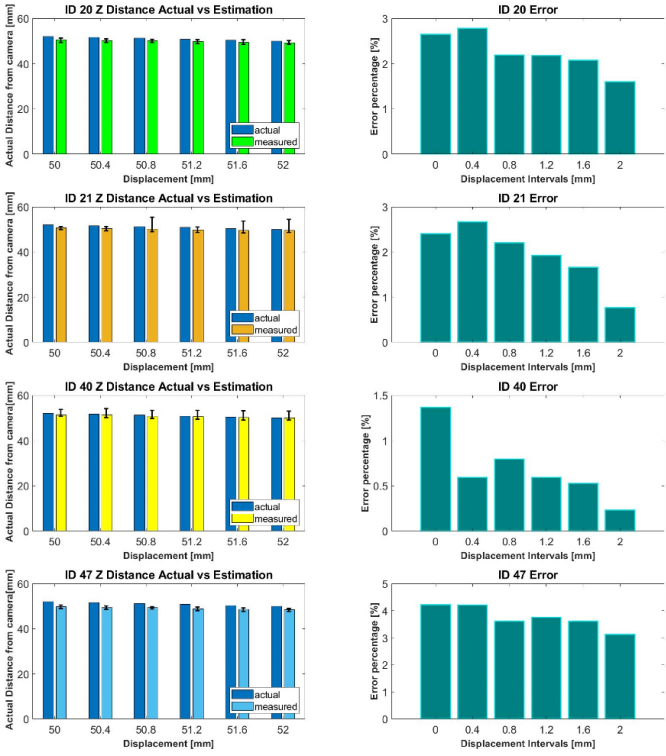

Fig. 4: Comparison of the Z depth estimation of exemplary AruCo Markers (i.e., ID 20, ID 21, ID 40, and ID 47 as marked in Fig. 2) with their actual displacement. The figure also shows the corresponding relative error percentages of these markers.

Figure 6 also shows the position of the exemplary markers (i.e., ID 12, ID 20, ID 21, ID 40, and ID 47 as marked in Fig. 2) color coded with respect to their X and Y position in the image space as the gel layer is sequentially deformed up to 2 mm with 0.4 mm intervals. As seen in this figure, QS-TS can identify and detect the markers in correct pattern sequences through the whole deformation procedure. In this figure, the deformation of each marker has been shown with a particular geometrical marker to better show its estimated deformation trajectory. Of note, as can be observed in Fig. 2 and expected, in the performed experiments, due to the deformation and dome-shape of the gel layer, each marker has experienced a lateral movement in the X and Y direction. Moreover, the calculated estimated distances ($d_E$) between different IDs greatly agree with their actual measured values ($d_A$), indicating the remarkable performance of the QS-

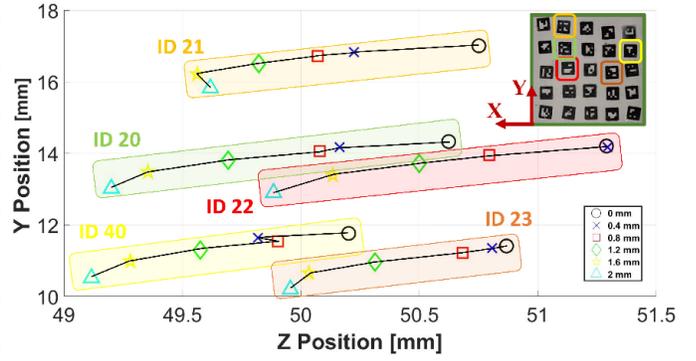

Fig. 5: Trajectories of exemplary ArUco markers (ID 20, ID 21, ID 22, ID 21, and ID 40) are demonstrated when the V-QTS has displaced a total of 0.2 mm with 0.4 mm intervals. Each marker is color-coded in order to identify their similar behavior easily. Each geometrical marker represents the position of ArUco markers during the deformation procedure.

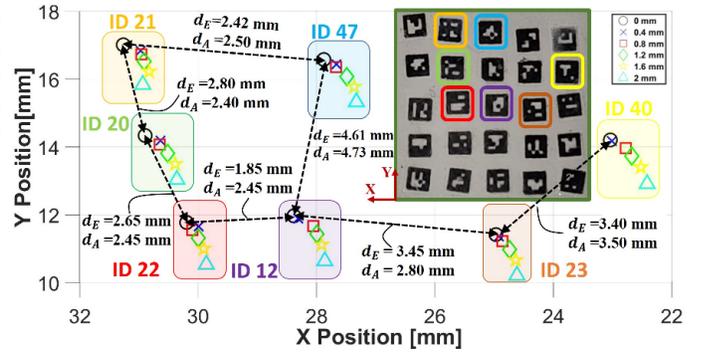

Fig. 6: Position of the exemplary markers (i.e., ID 12, ID 20, ID 21, ID 40, and ID 47 as marked in Fig. 2) color coded with respect to their X and Y position in the image space as the gel layer is sequentially deformed up to 2 mm with 0.4 mm intervals. The figure also compares the calculated estimated distances ($d_E$) between different IDs and their corresponding actual measured values ($d_A$).

TS in estimating the X and Y location of markers with respect to the camera location. As indicated, the error of estimated distances between all markers is less than 0.5 mm. Additionally, the deformation progression of each marker is consistent with the performed experiments.

## V. CONCLUSION

In this study, we presented the design, fabrication, and characterization of a novel ArUco markers-based Quantitative Surface Tactile Sensor (called QS-TS) to address the following limitations of conventional VTSs, including (i) the time-consuming and arduous fabrication methods for the marker attachment, and (ii) lack of a direct *quantitative* deformation evaluation of typical VTSs in real-time. Thanks to the use of ArUco markers, our novel proposed sensor, regardless of the placement of ArUco markers, enables an accurate estimation of the gel layer deformation in X, Y, and Z directions. The performance and efficacy of the proposed QS-TS in estimating the deformation of the sensor's gel layer were experimentally evaluated and verified. An accurate estimation of the deformation of the gel layer with a low relative error of < 5% in the Z direction and less than 0.5 mm in both the X and Y direction was achieved.

With this work, one can easily and quickly customize and fabricate high-resolution tactile sensors that suit the specific requirements of their robotic systems. Particularly, as the future direction of this study and similar to the literature (e.g., [23], [25]–[29]), we plan to integrate a modified version of QS-TS with a robotic hand to perform safe manipulation of objects and tactile servoing. Moreover, future work will include evaluating the sensor deformation estimation in the presence of shear and torsional deformation, using a better printing quality to minimize the size of the used ArUco markers and, therefore, improving the detection rate of the QS-TS.